\pdfoutput=1

\documentclass[11pt]{article}

\usepackage{acl}

\usepackage{times}
\usepackage{latexsym}
\usepackage[T1]{fontenc}

\usepackage[utf8]{inputenc}

\usepackage{microtype}

\usepackage{smile}
\usepackage{caption}
\usepackage{subcaption}
\usepackage{pifont}
\usepackage{enumitem}
\newcommand{\ours}{\text{SPADE}}
\newcommand{\xmark}{\ding{55}}

\title{Efficient Long Sequence Modeling via State Space Augmented Transformer}
     
\author{
Simiao Zuo\thanks{\hspace{4pt}Simiao Zuo and Xiaodong Liu made equal contributions. Work was done during Simiao Zuo's internship at Microsoft. Correspondence to \texttt{simiaozuo@gatech.edu} and \texttt{xiaodl@microsoft.com}.}$\hspace{4pt}^\ddagger$,
Xiaodong Liu$^*$\thanks{\hspace{4pt}Jian Jiao and Xiaodong Liu led the project.}$\hspace{4pt}^\diamond$,
Jian Jiao$^\dagger$$^\diamond$, Denis Charles$^\diamond$, Eren Manavoglu$^\diamond$, \\
\textbf{Tuo Zhao}$^\ddagger$ and \textbf{Jianfeng Gao$^\diamond$ } \\
$^\ddagger$Georgia Institute of Technology \ \
$^\diamond$Microsoft \\
}
\begin{document}
\maketitle

\begin{abstract}
Transformer models have achieved superior performance in various natural language processing tasks. However, the quadratic computational cost of the attention mechanism limits its practicality for long sequences. There are existing attention variants that improve the computational efficiency, but they have limited ability to effectively compute global information. In parallel to Transformer models, state space models (SSMs) are tailored for long sequences, but they are not flexible enough to capture complicated local information. We propose {\ours}, short for \underline{\textbf{S}}tate s\underline{\textbf{P}}ace \underline{\textbf{A}}ugmente\underline{\textbf{D}} Transform\underline{\textbf{E}}r. Specifically, we augment a SSM into the bottom layer of {\ours}, and we employ efficient local attention methods for the other layers. The SSM augments global information, which complements the lack of long-range dependency issue in local attention methods.
Experimental results on the Long Range Arena benchmark and language modeling tasks demonstrate the effectiveness of the proposed method. To further demonstrate the scalability of {\ours}, we pre-train large encoder-decoder models and present fine-tuning results on natural language understanding and natural language generation tasks.
\end{abstract}
\section{Introduction}

Transformer models have achieved superior performance on various natural language processing tasks such as language modeling \citep{dai2019transformer}, natural language generation \citep{brown2020language} and natural language understanding \citep{devlin2018bert, he2020deberta}.
These models leverage the attention mechanism \citep{vaswani2017attention}, which computes a dependency score for every pair of tokens in an input sequence. Therefore, full attention has a quadratic time and space complexity with respect to the sequence length. However, such a complexity is computationally prohibitive for tasks that involve long sequences, such as text summarization \citep{nallapati2016abstractive} and question answering \citep{kwiatkowski2019natural}. For example, empirically we find that a Transformer model (250M parameters) consumes over 80G of GPU memory when the sequence length is 8k.

Additionally, Transformer models equipped with the full attention are easy to overfit because of the lack of structural biases \citep{lin2022survey}.
That is, the attention mechanism does not assume any structural prior over the inputs.
For example, we even need order information (e.g., through sinusoidal encoding) to train a Transformer model.
Therefore, the full attention is too flexible such that Transformer models may easily overfit to the noise. This significantly limits the models' practicality in long sequence modeling, where the dependency signal is often weak and the signal-to-noise ratio is often low. 
Empirically, we find that on a two-way classification task, Transformer with the full attention has a 57.5\% accuracy, nearly 30\% less than state-of-the-art methods with powerful structural biases (see Section~\ref{sec:lra} for details).

Various approaches have been proposed to reduce the quadratic complexity and/or to introduce structural biases.
In \emph{approximation methods}, we approximate the full attention using fast algorithms with linear complexity. For example, we can approximate and speedup the computation of the attention score matrix (i.e., $\mathrm{softmax}(\Qb \Kb^\top/\sqrt{d})$ in Eq.~\ref{eq:attention}) using low-rank approximation \citep{wang2020linformer} or kernel methods \citep{peng2021random}.
However, even though these methods reduce the complexity of full attention, they inherit the lack of structural bias issue.

To incorporate structural biases to the Transformer model, \emph{partial attention} methods are proposed. Such methods can be further categorized into \emph{sparse attention} and \emph{clustering} methods. In sparse attention \citep{beltagy2020longformer}, each token only attends to a subset of all the tokens according to pre-defined sparsity patterns. In clustering methods \citep{kitaev2020reformer}, tokens are divided into several clusters, and only intra-cluster attention is performed. 
However, the introduced structural biases restrict the models' ability to capture global information. For example, in local-window attention, we assume each token only depends on its neighbors, such that we inevitably lose long-range and global information.

Contrary to partial attention, state space models (SSMs) introduce a different structural bias \citep{gu2021efficiently}, which is tailored for computing global information.
Specifically, SSMs design fixed global dependency patterns that facilitate effective and efficient computation.
These models can be seen as linear recurrent neural networks with specifically designed fixed weights. Moreover, efficient algorithms are crafted for training such models.
However, the integrated structural bias is restrictive in that SSMs are not refined enough to capture local information. This is because unlike attention, SSMs do not explicitly compute dependencies between input tokens.


We propose {\ours}, short for \underline{\textbf{S}}tate s\underline{\textbf{P}}ace \underline{\textbf{A}}ugmente\underline{\textbf{D}} Transform\underline{\textbf{E}}r. The proposed model is a multi-layer Transformer model that can effectively and efficiently capture complicated dependencies. Specifically, we augment a SSM into the bottom layer of the model, such that after this layer, inputs are integrated with global information. Because the SSM only provides coarse global information, at the subsequent top layers of {\ours}, we employ local attention variants to capture more complicated and refined local information.
In other words, in {\ours}, the SSM induces a strong structural bias that augments global information, and it complements the lack of long-range dependency issue in local attention methods.

We demonstrate the efficiency and effectiveness of {\ours} on various natural language processing tasks.
First, we show that the proposed method outperforms existing approaches on the Long Range Arena \citep{tay2020long} benchmark, which is designed to test models' ability in modeling long sequences.
Second, we show that in autoregressive language modeling, {\ours} is not only significantly faster than the vanilla Transformer \citep{vaswani2017attention}, but also yields better performance.
Third, we demonstrate the scalability of {\ours} by conducting language model pre-training and fine-tuning experiments. Specifically, we pre-train an encoder-decoder model similar to T5 \citep{raffel2020exploring}. And we fine-tune the model on various tasks, including natural language understanding and natural language generation benchmarks. In all the settings, {\ours} outperforms the baselines.
Finally, we provide analysis and ablation experiments to further demonstrate the effectiveness of the proposed method.

Our code\footnote{\url{https://github.com/microsoft/EfficientLongSequenceModeling}} and pre-trained model checkpoints\footnote{\url{https://github.com/namisan/mt-dnn}} are publicly available.
\section{Background}

\subsection{Attention Mechanism}


Suppose the input to the layer is $\Xb \in \RR^{L \times d}$, where $L$ is the sequence length and $d$ is the embedding dimension, then the attention mechanism outputs
\begin{align} \label{eq:attention}
    &\mathrm{Attn}(\Xb) = \mathrm{softmax}\left( \frac{\Qb \Kb^\top}{\sqrt{d}} \right) \Vb, \\
    &\text{where } \Qb = \Xb \Wb_q, \ \Kb = \Xb \Wb_k, \ \Vb = \Xb \Wb_v. \notag
\end{align}
Here $\Wb_q, \Wb_k, \Wb_v \in \RR^{d\times d}$ are learnable weights. The attention mechanism can simultaneously compute the alignment between any pair of input tokens, such that it models long-range dependencies better than recurrent neural networks. Specifically, denote the attention score matrix $\Ab = \mathrm{softmax}(\Qb \Kb^\top / \sqrt{d}) \in \RR^{L \times L}$. Then, $\Ab_{ij}$ captures the alignment between the $i$-th and the $j$-th input tokens.


\subsection{State Space Models}

\textbf{Continuous time state space model.}
A continuous time latent space model maps a $1$-dimensional input signal $u(t)$ to a $d_s$-dimensional latent state $x(t)$, after which $x(t)$ is mapped to a $1$-dimensional output signal $y(t)$. Concretely,
\begin{align} \label{eq:ssm}
    x'(t) = \Ab x(t) + \Bb u(t), \quad y(t) = \Cb x(t).
\end{align}
Here, $\Ab \in \RR^{d_s \times d_s}$, $\Bb \in \RR^{d_s}$ and $\Cb \in \RR^{d_s}$.

Existing works leverage Eq.~\ref{eq:ssm} to model long sequences. For example, \citet{gu2020hippo} claim that randomly initialized parameters $\Ab$, $\Bb$ and $\Cb$ cannot model long-range dependencies well. Subsequently, a class of matrices (termed HiPPO, high-order polynomial projection operators) are proposed to initialize $\Ab$. The HiPPO matrices are designed such that the state $x(t)$ at time $t$ can memorize the history of the input $u(t)$ up to time $t$.


\vspace{0.05in} \noindent
\textbf{Discrete time state space model.}
In practice, we often work with discrete sequences such as natural language inputs $(u_0, u_1, \cdots, u_L)$, where $L$ is the sequence length. To facilitate modeling discrete data, the model in Eq.~\ref{eq:ssm} can be discretized (using the bilinear method) by a step size $\Delta$, such that
\begin{align} \label{eq:ssm-discrete}
    &x_k = \overline{\Ab} x_{k-1} + \overline{\Bb} u_{k}, \quad y_k = \overline{\Cb} x_k, \\
    &\text{where }\overline{\Ab} = (\Ib - \Delta/2 \cdot \Ab)^{-1} (\Ib + \Delta/2 \cdot \Ab), \notag \\ 
    &\quad\quad\ \ \ \overline{\Bb} = (\Ib - \Delta/2 \cdot \Ab)^{-1} \Delta \Bb, \quad \overline{\Cb} = \Cb. \notag
\end{align}
We unroll the above recurrent representation, after which we have
\begin{align*}
    y_k = \overline{\Cb} \overline{\Ab}^k \overline{\Bb} u_0 + \cdots + \overline{\Cb} \overline{\Ab} \overline{\Bb} u_{k-1} + \overline{\Cb} \overline{\Bb} u_k.
\end{align*}
This can be written as a convolutional representation $y = \overline{\Kb} * u$, where the convolution kernel
\begin{align} \label{eq:ssm-conv}
    &\overline{\Kb} \in \RR^L = \left( \overline{\Cb} \overline{\Bb}, \overline{\Cb} \overline{\Ab} \overline{\Bb}, \cdots, \overline{\Cb} \overline{\Ab}^{L-1} \overline{\Bb} \right).
\end{align}
Here, ``$*$'' is the discrete convolution operator, $u$ represents the input sequence $(u_0, u_1, \cdots, u_L)$, and $y$ represents the corresponding output sequence $(y_0, y_1, \cdots, y_L)$.

In Eq.~\ref{eq:ssm-conv}, the output $y$ can be computed efficiently given that the convolution kernel $\overline{K}$ is known. However, computing the kernel is non-trivial. Most of existing algorithms have $O(L^2)$ time and space complexity.

\vspace{0.05in} \noindent
\textbf{Structured State Space Sequence model (S4).}
\citet{gu2021efficiently} develop the S4 model to efficiently compute Eq.~\ref{eq:ssm-conv}. Specifically, $\Cb$ in Eq.~\ref{eq:ssm} is randomly initialized, and $\Ab$ and $\Bb$ are initialized as
\begin{align} \label{eq:s4-init}
    &\Ab = \Ab^{(d_s)} - \Pb \Pb^\top, \quad \Bb_i = (2i+1)^{\frac{1}{2}}, \\
    &\text{where } \Pb_i = \left( i+1/2 \right)^{1/2}, \notag \\
    &\Ab^{(d_s)}_{ij} = -
    \begin{cases}
        (i+\frac{1}{2})^{\frac{1}{2}} (j+\frac{1}{2})^{\frac{1}{2}}, & i > j, \\
        \frac{1}{2}, & i = j, \\
        -(i+\frac{1}{2})^{\frac{1}{2}} (j+\frac{1}{2})^{\frac{1}{2}}, & i < j. \\
    \end{cases} \notag
\end{align}
Subsequently, the convolution kernel $\overline{\Kb}$ in Eq.~\ref{eq:ssm-conv} can be computed efficiently with $O(L)$ time and space complexity.
Subsequently, for an input $u$, the S4 output $y=\overline{\Kb} * u$ can be computed efficiently.

\section{Method}

We first conduct experiments to demonstrate that SSMs do not model local information well. Then, we present {\ours}, which effectively combines global and local information by augmenting SSMs into the Transformer architecture.

\subsection{Attention vs. State Space Models}

\begin{figure}[t!]
    \centering
    \includegraphics[width=0.45\linewidth]{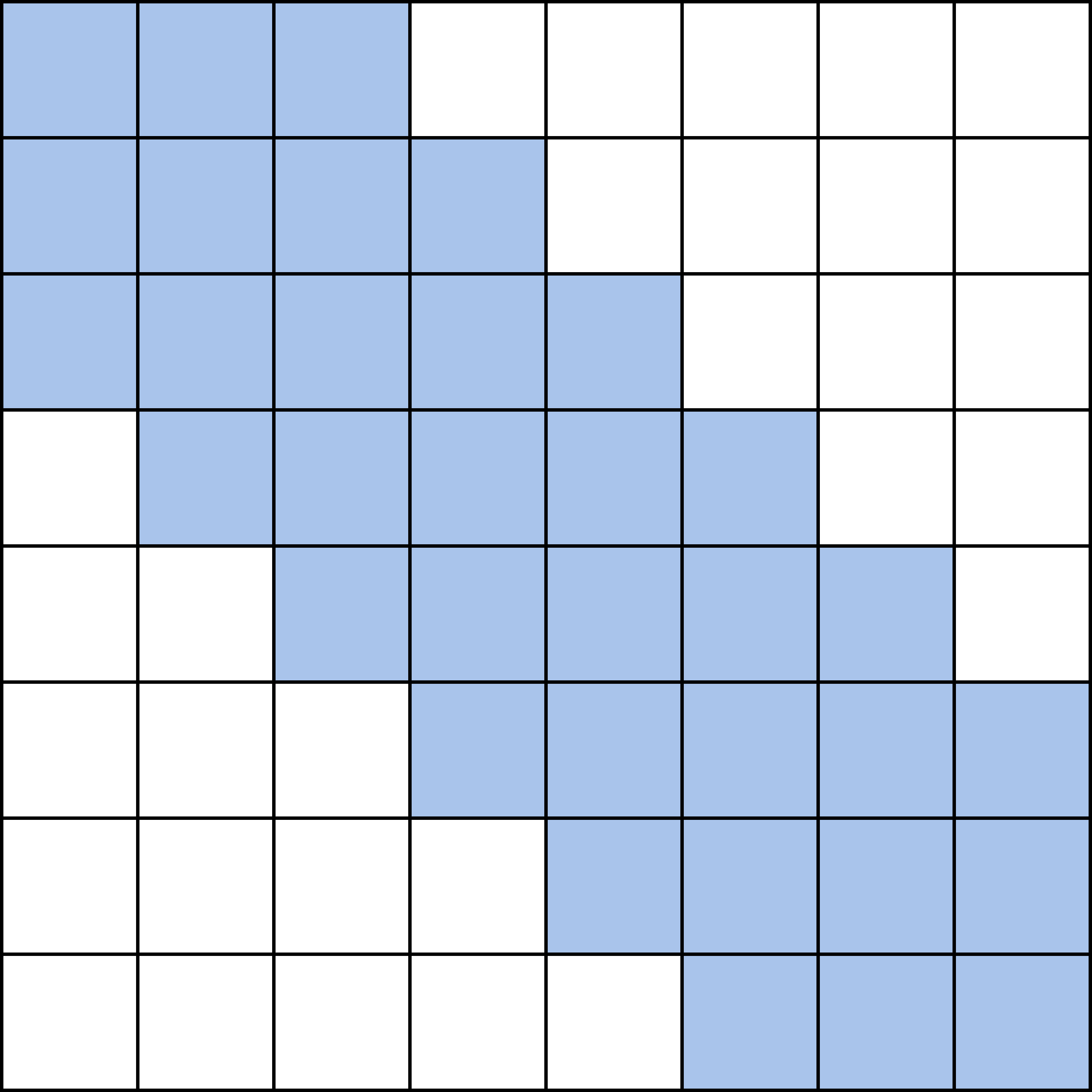} \hfill
    \includegraphics[width=0.45\linewidth]{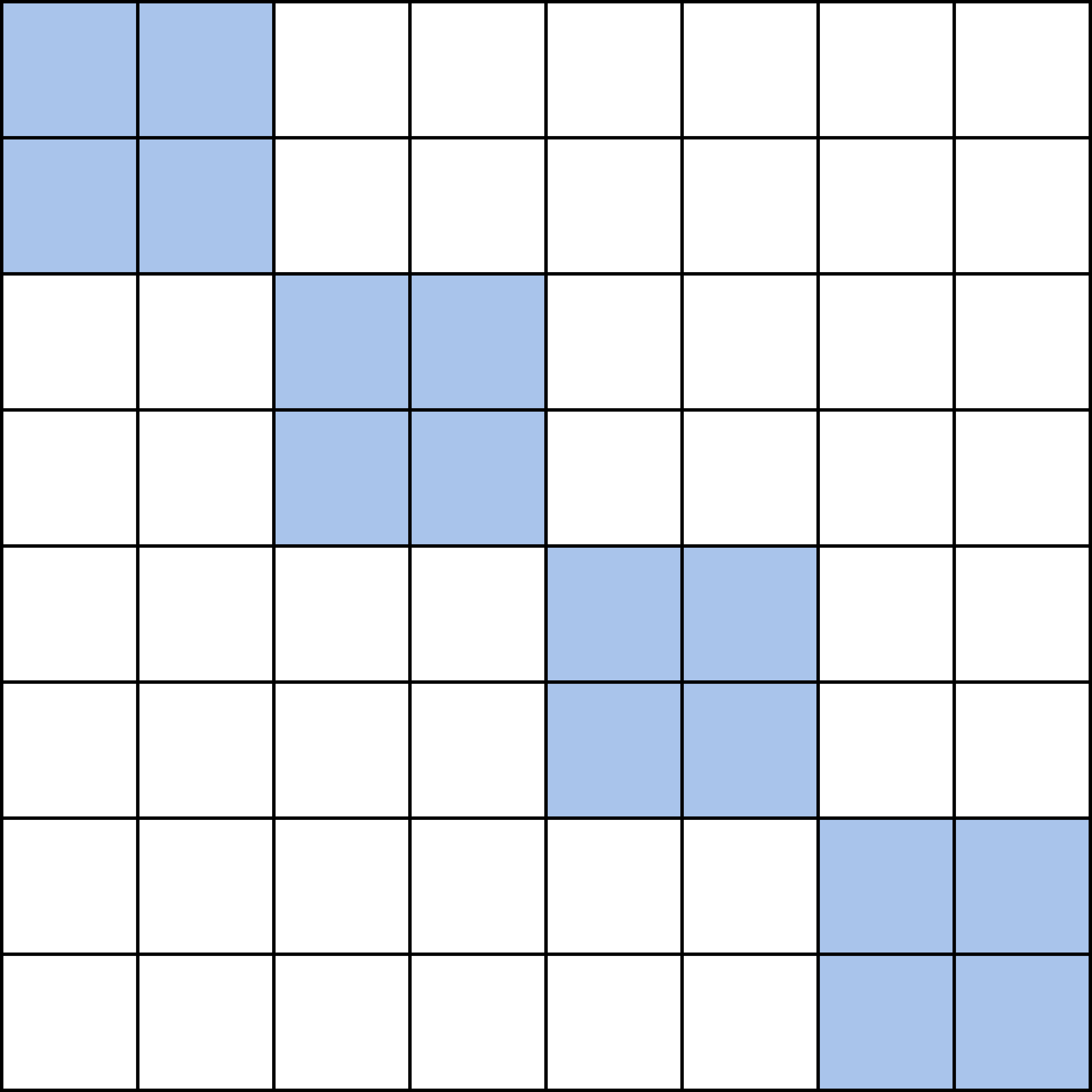}
    \vspace{-0.05in}
    \caption{Illustration of window attention (left) and chunk attention (right). For window attention, the window size is 2 (on each side); for chunk attention, the chunk size is 2.}
    \label{fig:attention}
\end{figure}

The motivation behind {\ours} is that even though SSMs perform well on several long sequence classification tasks \citep{gu2021efficiently}, they perform poorly on language modeling, which is a fundamental task in natural language processing.
To demonstrate such an observation, we compare S4 with Transformer with full attention and Transformer with local (window and chunk) attention. In local attention, each token can only attend to its neighboring tokens (see Figure \ref{fig:attention} for illustrations).
We conduct experiments on token-level language modeling. In this setting, local information is more important than global information. This is because in practice, we rarely see words (tokens) that are thousands of positions apart exhibit strong dependencies \citep{sukhbaatar2019adaptive}.

\begin{figure}[h!]
    \centering
    \includegraphics[width=0.7\linewidth]{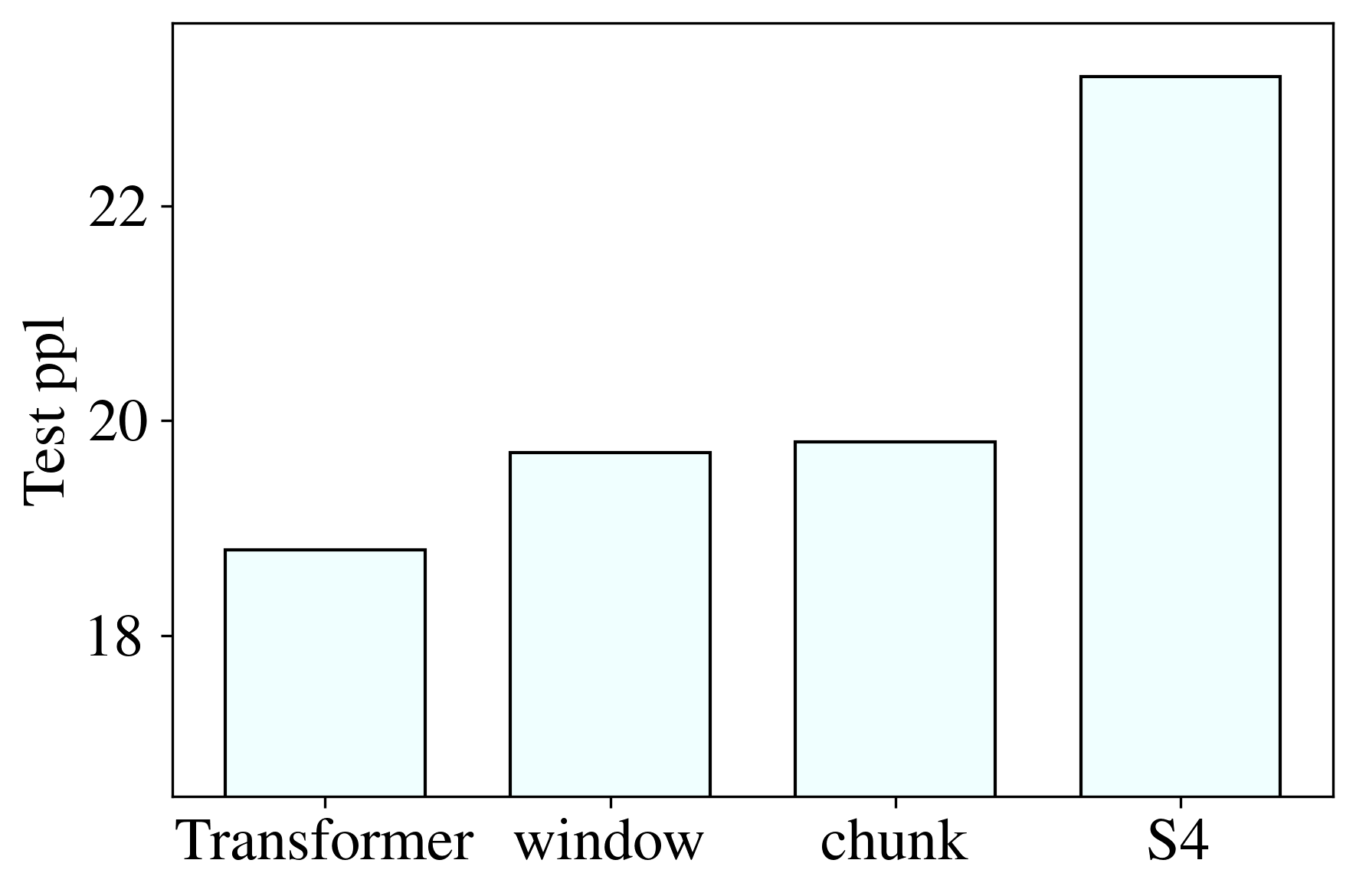}
    \vspace{-0.1in}
    \caption{Performance of Transformer with full attention, window attention, chunk attention, and S4. We conduct language modeling experiments (see Section~\ref{sec:lm}), and the sequence length is 3k.}
    \label{fig:compare}
\end{figure}

Experimental results are illustrated in Figure~\ref{fig:compare}. We see that both Transformer with full attention and Transformer with local attention (e.g., window and chunk) outperforms S4.
Notice that replacing full attention with local attention does not significantly hurt model performance, indicating that local information is more important in this setting.
We remark that SSMs such as S4 produces a fixed dependency pattern, e.g., the convolution kernel in Eq.~\ref{eq:ssm-conv}. Moreover, unlike the attention mechanism, SSMs do not explicitly compute dependencies between tokens. Therefore, SSMs are not refined enough to capture local information, such that they perform poorly on language modeling tasks.


\subsection{{\ours}: State Space Augmented Transformer}

\begin{figure}
    \centering
    \includegraphics[width=0.39\linewidth]{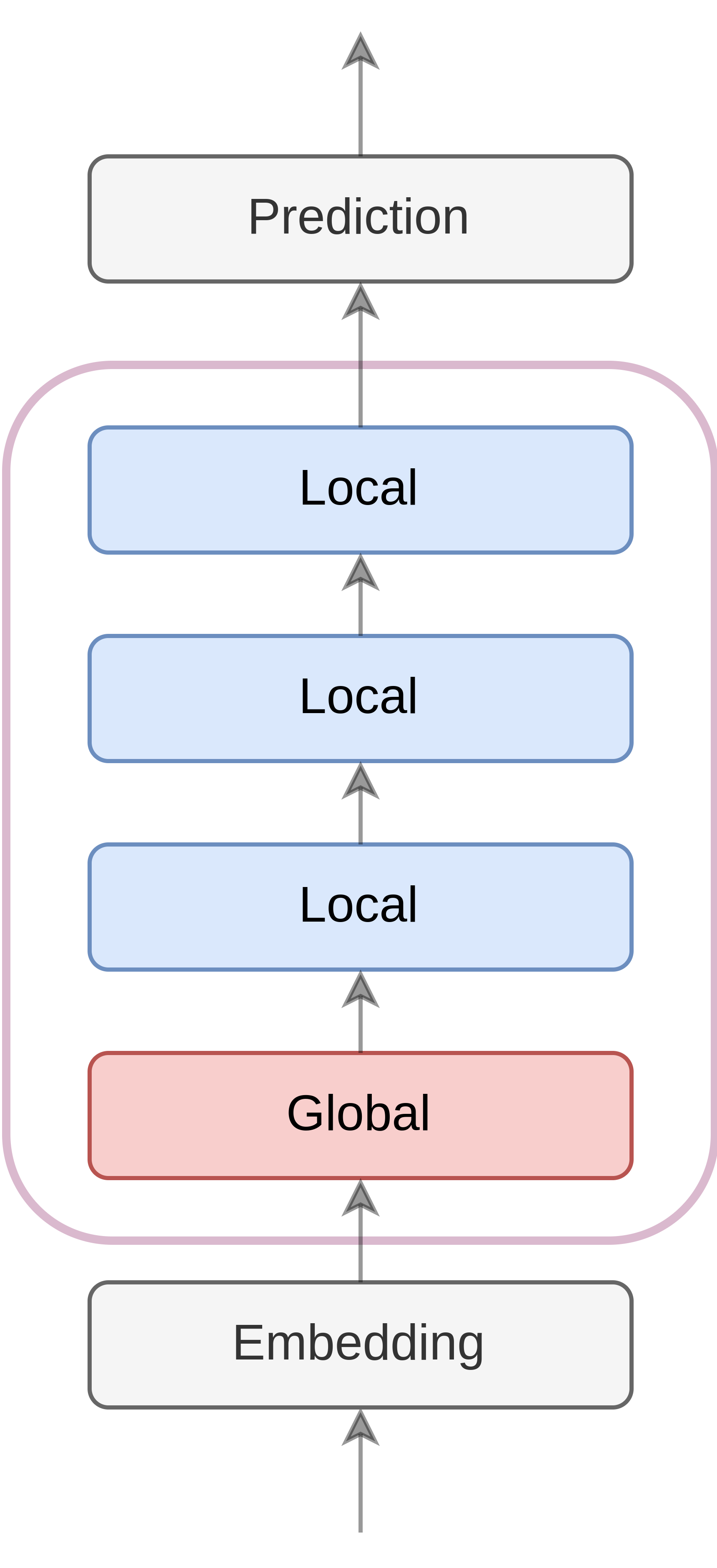} \hfill
    \includegraphics[width=0.51\linewidth]{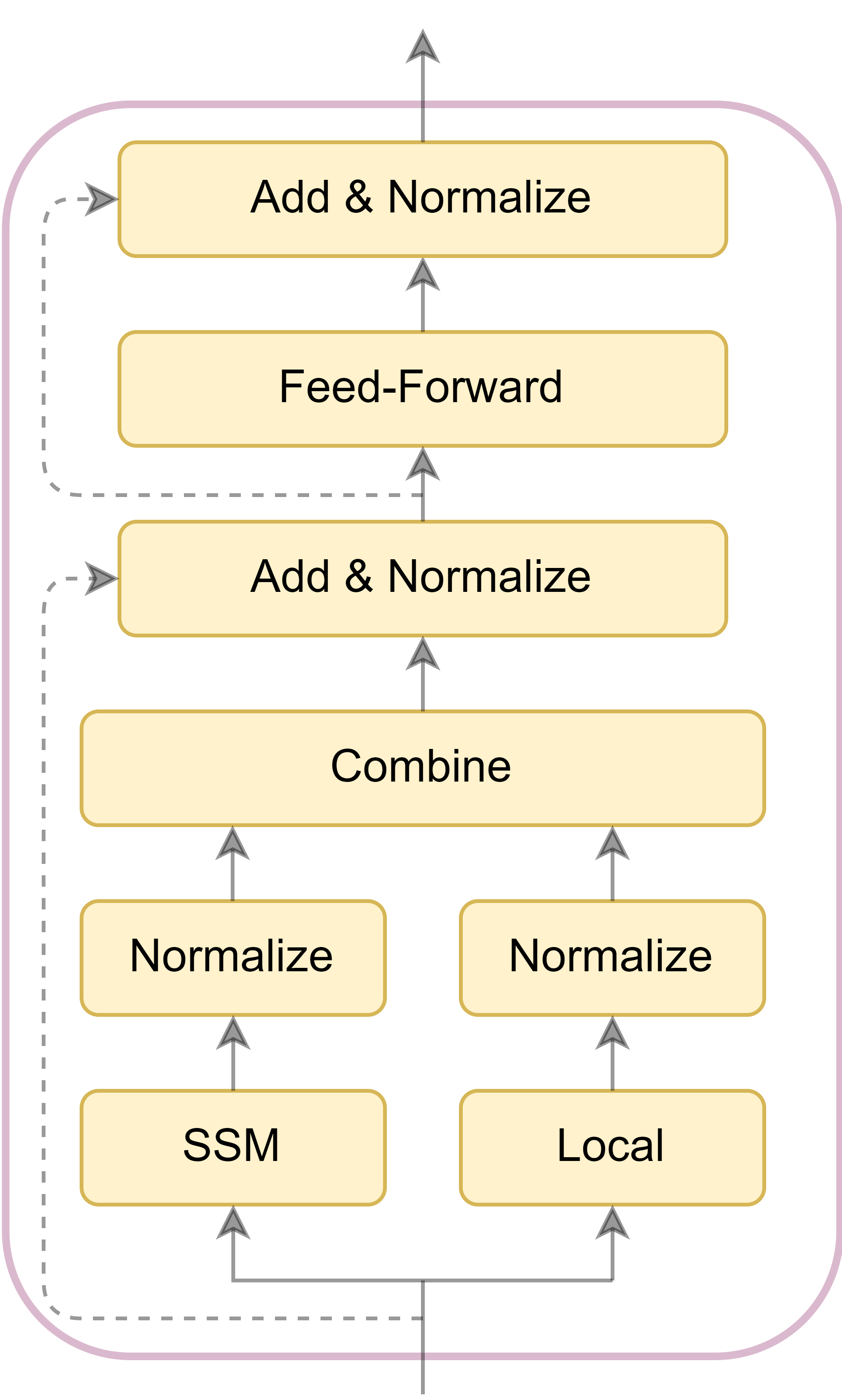}
    \vspace{-0.05in}
    \caption{Demonstration of {\ours} with 4 layers. Left: model overview; Right: details of the global layer.}
    \label{fig:model-arch}
\end{figure}

We propose {\ours}, which is a multi-layer Transformer model that can capture complicated global and local information.
The overall architecture of {\ours} is shown in Figure~\ref{fig:model-arch} (left).
The proposed model employs a hierarchical structure. Specifically, at the bottom layer of {\ours} (termed the \emph{global} layer), we capture global dependencies using a SSM. Because the SSM only provides coarse global information, the subsequent \emph{local} layers facilitate the model to handle more refined and complicated local dependencies. In other words, the SSM induces a strong structural bias that augments global information to the inputs.

To instantiate the local layer, we replace the full attention in the conventional Transformer layer with off-the-shelf efficient local attention methods. {\ours} is flexible to accommodate different approaches, such as window attention and chunk attention (see Figure~\ref{fig:attention} for illustrations).

In the global layer (Figure~\ref{fig:model-arch}, right), given the input $\Xb$ to the layer, we have the output $\Yb$ as
\begin{align*}
    &\Xb_{\text{local}} = \mathrm{Local} \left( \mathrm{LN}(\Xb) \right), \\
    &\Xb_{\text{global}} = \mathrm{SSM} \left( \mathrm{LN}(\Xb) \right), \\
    &\Xb_a = \Wb \left[ \mathrm{LN}(\Xb_{\text{local}}), \mathrm{LN}(\Xb_{\text{global}}) \right] + \Xb, \\
    &\Yb = \mathrm{FFN} \left( \mathrm{LN}(\Xb_a) \right) + \Xb_a.
\end{align*}
Here, $\mathrm{LN}(\cdot)$ denotes layer normalization \citep{ba2016layer}, $\mathrm{FFN}(\cdot)$ denotes a two-layer feed-forward neural network, and $\Wb$ is a trainable weight that combines local and global representations. Notice that we apply normalization to $\Xb_{\text{local}}$ and $\Xb_{\text{global}}$ to align their scales. In this work, we choose S4 as the state space model.

We remark that because of the sequential nature of SSMs (Eq.~\ref{eq:ssm-discrete}), the global layer can encode positional information of the inputs. Therefore, we do not need additional fixed-length positional embedding techniques \citep{devlin2018bert}.
Such a property enables {\ours} to extrapolate to longer sequence length during testing, e.g., we can train a model with sequence length 512 and test the model with sequence length 1k.

\section{Experiments}

In the experiments, we implement all the models using \textit{PyTorch} \citep{paszke2019pytorch} and \textit{Fairseq} \citep{ott2019fairseq}. Training details such as hyper-parameter settings are deferred to the appendix.

\subsection{Long Range Arena}
\label{sec:lra}

\noindent
\textbf{Dataset.}
We evaluate the effectiveness of the proposed model on Long Range Arena (LRA, \citealt{tay2020long}), which is a benchmark tailored for evaluating models' ability in modeling long sequences. The benchmark contains six tasks:
ListOps, which tests the capability of modeling hierarchically structured data \citep{nangia2018listops};
byte-level text classification on the IMDB movie review dataset (Text, \citealt{maas2011learning});
byte-level document retrieval on the ACL anthology network (Retrieval, \citealt{radev2013acl});
pixel-level image classification on CIFAR-10 (Image, \citealt{krizhevsky2009learning});
Pathfinder, which tests the capability in modeling spatial dependency \citep{linsley2018learning};
and a longer version of Pathfinder (Path-X, \citealt{tay2020long}).

\begin{table*}[t!]
\centering \small
\resizebox{1.0\textwidth}{!}{
\begin{tabular}{l|cccccc|c}
\toprule
Dataset & \textbf{Listops} & \textbf{Text} & \textbf{Retrieval} & \textbf{Image} & \textbf{Pathfinder} & \textbf{Path-X} & \textbf{Avg.} \\ \midrule
Sequence length & 2k & 4k & 8k & 1k & 1k & 16k & --- \\ \midrule
Random & 10.00 & 50.00 & 50.00 & 10.00 & 50.00 & 50.00 & 36.67 \\
Transformer (full) \citep{vaswani2017attention} & 36.37 & 64.27 & 57.46 & 42.44 & 71.40 & \xmark & 53.66 \\
Transformer (window) & 15.82 & 52.98 & 53.39 & 41.46 & 66.63 & \xmark & 46.71 \\
Sparse Trans. \citep{child2019generating} & 17.07 & 63.58 & 59.59 & 44.24 & 71.71 & \xmark & 51.03 \\
Longformer \citep{beltagy2020longformer} & 35.63 & 62.85 & 56.89 & 42.22 & 69.71 & \xmark & 52.88 \\
Linformer \citep{wang2020linformer} & 35.70 & 53.94 & 52.27 & 38.56 & 76.34 & \xmark & 51.14 \\
Reformer \citep{kitaev2020reformer} & 37.27 & 56.10 & 53.40 & 38.07 & 68.50 & \xmark & 50.56 \\
Sinkhorn Trans. \citep{tay2020sparse} & 33.67 & 61.20 & 53.83 & 41.23 & 67.45 & \xmark & 51.23 \\
Synthesizer \citep{tay2021synthesizer} & 36.99 & 61.68 & 54.67 & 41.61 & 69.45 & \xmark & 52.40 \\
BigBird \citep{zaheer2020big} & 36.05 & 64.02 & 59.29 & 40.83 & 74.87 & \xmark & 54.17 \\
Linear Trans. \citep{katharopoulos2020transformers} & 16.13 & 65.90 & 53.09 & 42.34 & 75.30 & \xmark & 50.46 \\
Performer \citep{choromanski2020rethinking} & 18.01 & 65.40 & 53.82 & 42.77 & 77.05 & \xmark & 51.18 \\
FNet \citep{lee2021fnet} & 35.33 & 65.11 & 59.61 & 38.67 & 77.80 & \xmark & 54.42 \\
Nystromformer \citep{xiong2021nystromformer} & 37.15 & 65.52 & 79.56 & 41.58 & 70.94 & \xmark & 57.46 \\
Luna-256 \citep{ma2021luna} & 37.25 & 64.57 & 79.29 & 47.38 & 77.72 & \xmark & 59.37 \\
FMMformer \citep{nguyen2021fmmformer} & 36.74 & 67.84 & 81.88 & 45.10 & 72.12 & \xmark & 60.74 \\
S4 \citep{gu2021efficiently} & 58.35 & 76.02 & 87.09 & 87.26 & 86.05 & 88.10 & 80.48 \\
MEGA-chunk \citep{ma2022mega} & 58.76 & 90.19 & 90.97 & 85.80 & 94.41 & 93.81 & 85.66 \\
\midrule
{\ours} (softmax-window) & 59.70 & 87.55 & 90.13 & \textbf{89.11} & \textbf{96.42} & 94.22 & 86.19 \\
{\ours} (MEGA-chunk) & \textbf{60.50} & \textbf{90.69} & \textbf{91.17} & 88.22 & 96.23 & \textbf{97.60} & \textbf{87.40} \\
\bottomrule
\end{tabular}
}
\vspace{-0.05in}
\caption{Experimental results on Long Range Arena (LRA). Path-X uses 16k as the input sequence length, and ``\xmark'' indicates unavailable results due to computational constraints. All the baseline results, except for MEGA-chunk, are from \citet{gu2021efficiently}. MEGA-chunk results are from \citet{ma2022mega}.}
\label{tab:lra-results}
\end{table*}

\vspace{0.05in} \noindent
\textbf{Models.}
Following the standard setting \citep{tay2020long}, we use small models (less than 2M parameters) for all the tasks. We limit the computational budget such that all the models are trained with similar speed for the same amount of time.

To aggregate local information, we consider two approaches: window attention and chunk attention.
For window attention, we sparsify the conventional softmax attention (termed \textit{softmax-window}); and for chunk attention, we sparsify MEGA \citep{ma2022mega}, which employs a gated attention technique (termed \textit{MEGA-chunk}).
For window attention, we set the window size to 128, except Path-X, where we set the window size to 1024. For chunk attention, we set the chunk size to 128, except Path-X, where we set the chunk size to 4096.

\vspace{0.05in} \noindent
\textbf{Results.}
Experimental results are summarized in Table~\ref{tab:lra-results}. We see that both variants of {\ours} (softmax-window and MEGA-chunk) significantly outperform all the baselines in terms of average accuracy. For example, the window attention variant outperforms the best-performing baseline (MEGA-chunk) by 0.5\%, and the chunk attention variant has a 1.8\% performance gain. Therefore, {\ours} is more suitable to model long sequences than existing approaches.

\subsection{Language Modeling}
\label{sec:lm}

We further evaluate our model by conducting language modeling experiments on Wikitext-103. The dataset contains English-language Wikipedia articles, and the total number of tokens is 103M.
In all the experiments, we follow the settings in \citet{baevski2018adaptive}, where we use a large-scale Transformer model with 16 layers and about 250M parameters. We set the input sequence length to 3k and train for 286k steps.
Similar to the LRA experiments, we equip {\ours} with either window attention (softmax-window) or chunk attention (MEGA-chunk). Additionally, we evaluate another efficient attention variant: FLASH-chunk, where we sparsify FLASH \citep{hua2022transformer}, a gated attention method similar to MEGA. 

\begin{table}[t!]
\centering \small
\begin{tabular}{l|c}
\toprule
& Test ppl \\ \midrule
Transformer \citep{vaswani2017attention} & 18.8 \\ \midrule
Transformer (window) & 19.7 \\
S4 \citep{gu2021efficiently} & 23.2 \\ 
FLASH-chunk \citep{hua2022transformer} & 20.9 \\
MEGA-chunk \citep{ma2022mega} & 19.8 \\ \midrule
{\ours} (FLASH-chunk) & 19.9 \\
{\ours} (MEGA-chunk) & 19.5 \\
{\ours} (softmax-window) & \textbf{18.5} \\
\bottomrule
\end{tabular}
\vspace{-0.05in}
\caption{Experimental results on Wikitext-103. For window attention (softmax-window), we set the window size to 512; for chunk attention (FLASH and MEGA), we set the chunk size to 512.}
\label{tab:wikitext-results}
\end{table}

\begin{table*}[t!]
\centering \small
\resizebox{1.0\textwidth}{!}{
\begin{tabular}{l|cccccccc|c}
\toprule
& \textbf{RTE} & \textbf{MRPC} & \textbf{CoLA} & \textbf{SST-2} & \textbf{STS-B} & \textbf{QNLI} & \textbf{QQP} & \textbf{MNLI-m/mm} & \textbf{Avg.} \\
& Acc & Acc/F1 & Mcc & Acc & P/S Corr & Acc & Acc/F1 & Acc & \textbf{Score} \\ \midrule
T5\textsubscript{base} \citep{raffel2020exploring} & 76.9 & 90.8/-- & 55.5 & 92.8 & 86.5 & 91.9 & 90.9/-- & 84.4/83.5 & --- \\
T5\textsubscript{base} (re-imp) & 78.0 & 91.7/88.6 & 61.5 & 93.6 & 88.2 & 92.9 & 91.2/87.9 & 87.0/86.9 & 85.1 \\ \midrule
{\ours}\textsubscript{base} & 77.9 & 92.2/89.0 & 63.2 & 94.0 & 87.9 & 92.8 & 91.6/88.2 & 87.1/87.2 & 85.4 \\
{\ours}\textsubscript{base++} & \textbf{80.5} & \textbf{92.3}/\textbf{89.2} & \textbf{64.7} & \textbf{95.9} & \textbf{89.2} & \textbf{93.9} & \textbf{91.7}/\textbf{88.4} & \textbf{89.6}/\textbf{89.2} & \textbf{86.8} \\
\bottomrule
\end{tabular}
}
\vspace{-0.05in}
\caption{Experimental results on GLUE development set. ``T5\textsubscript{base}'' results are from \citealt{raffel2020exploring}. For   ``T5\textsubscript{base} (re-imp)'', we re-implement the T5 pre-training procedures, and we fine-tune the re-implemented T5 model.}
\label{tab:glue-t5}
\end{table*}

Experimental results are presented in Table~\ref{tab:wikitext-results}.
From the results, we see that by combining global and local information, the proposed model achieves significant performance improvement and outperform all the baselines. For example, the vanilla window attention has a 19.7 perplexity on the test set, and by integrating a SSM into {\ours}, we achieve a 1.2 perplexity gain.
We remark that {\ours} with softmax-window is not only significantly faster than the Transformer with full attention, but also yields a better performance.

\noindent\textbf{Remark.} We remark that we do not need to train the S4 in the bottom layer of {\ours} to achieve the performance in Table~\ref{tab:wikitext-results}. That is, we initialize the parameters in S4 using Eq.~\ref{eq:s4-init}, and the parameters are frozen during training. This is because even without training, the initialization of S4 yields intriguing theoretical properties, which facilitates S4's ability to capture global information.

\section{Language Model Pre-Training}

We implement model pre-training using \textit{Fairseq}, and we implement model fine-tuning using \textit{MT-DNN} \citep{liu2019multi, liu202mtdnnt}. Note that all our experiments only use single task fine-tuning. Details such as hyper-parameter settings are deferred to the appendix.

\subsection{Pre-Training Details}

To demonstrate the scalability of the proposed method, we pre-train an encoder-decoder variant of {\ours}. The model architecture is the same as T5\textsubscript{base} \citep{raffel2020exploring}, except that we use post-layernorm instead of pre-layernorm to improve model performance \cite{liu2020admin, xiong2020layer}. The embedding dimension is 768, the hidden dimension of the FFN is 3072, the number of attention heads is 12, and both the encoder and the decoder have 12 layers. 
We add a S4 module to the bottom layer of {\ours}, and the parameters of the S4 are fixed after initialization (Eq.~\ref{eq:s4-init}). We use the softmax-window attention as the local information extractor, where we set the window size to 128. The model contains about 290M parameters.

We consider two pre-training settings with different datasets and number of training steps:
\begin{itemize}[leftmargin=*]
    \item[$\diamond$] {\ours}\textsubscript{base}: We follow the pre-training settings in BERT \citep{devlin2018bert}. Specifically, we train the model on Wikipedia \citep{devlin2018bert} and BookCorpus \citep{zhu2015aligning}. We set the sequence length to 1024 and the batch size to 2048. We train the model for 125k steps.
    \item[$\diamond$] {\ours}\textsubscript{base++}: We follow the pre-training settings in RoBERTa \citep{liu2019roberta}. Specifically, we train the model on Wikipedia \citep{devlin2018bert}, BookCorpus \citep{zhu2015aligning}, STORIES \citep{trinh2018simple}, CC-News \citep{liu2019roberta}, and OpenWebText \citep{gokaslan2019openwebtext}. We train the model for 2M steps, with sequence length 1024 and batch size 2048.
\end{itemize}

We remark that because the S4 module is not trained after proper initialization, and we do not use fixed-length positional embedding, our pre-trained model can extrapolate to any sequence length. For example, we can set the sequence length to 2k during fine-tuning, which is longer than the sequence length used in pre-training.

\newcolumntype{C}{@{\hskip4pt}c@{\hskip4pt}}
\begin{table*}[t!]
\centering \small
\begin{tabular}{@{\hskip4pt}l@{\hskip4pt}|CCC|CCCC}
\toprule
& \# Train & \# Validation & \# Test & Mean & Median & Max & 90th percentile \\ \midrule
arXiv \citep{cohan2018discourse} & 203,037 & 6,436 & 6,440 & 10,720 & 8,519 & 378,825 & 20,170 \\
CNN/DailyMail \citep{nallapati2016abstractive} & 287,113 & 13,368 & 11,490 & 982 & 894 & 5,268 & 1,659 \\
MediaSum \citep{zhu2021mediasum} & 443,596 & 10,000 & 10,000 & 2,302 & 1,748 & 125,974 & 4,128 \\
MultiNews \citep{fabbri2019multi} & 44,972 & 5,622 & 5,622 & 2,594 & 1,902.5 & 683,544 & 4,853 \\
\bottomrule
\end{tabular}
\vspace{-0.05in}
\caption{Statistics and sources of abstractive summarization datasets.}
\label{tab:summarization-dataset}
\end{table*}

\subsection{Natural Language Understanding}

We fine-tune the pre-trained models on the General Language Understanding Evaluation (GLUE) benchmark \cite{wang2018glue}, which is a collection of natural language understanding tasks.
The benchmark includes two single-sentence classification tasks: CoLA \citep{cola2018} is a linguistic acceptability task; and SST-2 \citep{sst2013} is a binary classification task that classifies movie reviews to positive or negative. 
The benchmark also contains three similarity and paraphrase tasks: STS-B \citep{sts-b2017} is a text similarity task; MRPC \citep{mrpc2005} is a paraphrase detection task; and QQP is a duplication detection task.
Additionally, there are natural language inference tasks: MNLI \citep{mnli2018}; QNLI \citep{squad1}; RTE \citep{rte1, rte2, rte3, rte5}.
Dataset details are summarized in Appendix~\ref{app:dataset}.

We do not consider the long sequence setting in these tasks. All models (T5 and {\ours}) are fine-tuned under the sequence length of 512.

Experimental results are presented in Table~\ref{tab:glue-t5}. We see that {\ours}\textsubscript{base++} significantly outperforms T5\textsubscript{base}. For example, T5\textsubscript{base} has a 85.1 average score, and our model outperforms it by 1.7 average score. Recall that the sequence length is set to 512, which is the standard setting instead of the long-sequence setting. Therefore, the results indicate that {\ours} is universal in that it is suitable to model both long and short sequences.

\subsection{Natural Language Generation}

We also fine-tune the pre-trained models on several abstractive summarization datasets. The sources and statistics are summarized in Table~\ref{tab:summarization-dataset}.
We use ROUGE-2 scores as the evaluation metric.

We compare {\ours} with LongT5 \citep{guo2021longt5}, which is a T5 variant that is tailored for long sequences. Note that LongT5 uses PEGASUS-style \citep{zhang2020pegasus} Principle Sentences Generation as its pre-training objective, such that LongT5 is tailored for natural language generation tasks (e.g., summarization). In contrast, {\ours} is a more general model in that it also has superior performance on natural language understanding tasks. We remark that LongT5 uses C4 \citep{raffel2020exploring} as its pre-training dataset, which is much larger and diverse than the dataset we adopted.

\begin{table}[t!]
\centering \small
\begin{tabular}{l|cc|cc}
\toprule
& \multicolumn{2}{c|}{\textbf{arXiv}} & \multicolumn{2}{c}{\textbf{CNN/DailyMail}} \\
& Length & R-2 & Length & R-2 \\ \midrule
LongT5\textsubscript{base} & 4k & 18.54 & 4k & 20.11 \\
LongT5\textsubscript{large} & 16k & 21.63 & 4k & 20.51 \\
LongT5\textsubscript{xl} & 16k & 21.92 & 4k & 21.40 \\ \midrule
{\ours}\textsubscript{base++} & 16k & 21.65 & 4k & 20.40 \\
\toprule
& \multicolumn{2}{c|}{\textbf{MediaSum}} & \multicolumn{2}{c}{\textbf{MultiNews}} \\
& Length & R-2 & Length & R-2 \\ \midrule
LongT5\textsubscript{base} & 4k & 18.35 & 4k & 17.37 \\
LongT5\textsubscript{large} & 4k & 19.04 & 8k & 18.44 \\
LongT5\textsubscript{xl} & 4k & 19.66 & 8k & 19.43 \\ \midrule
{\ours}\textsubscript{base++} & 4k & 19.03 & 8k & 19.63 \\
\bottomrule
\end{tabular}
\vspace{-0.05in}
\caption{Experimental results (ROUGE-2) on test sets. LongT5 results are from \citealt{guo2021longt5}.}
\label{tab:summarization-results}
\end{table}

Experimental results are summarized in Table~\ref{tab:summarization-results}. From the results, we see that our model significantly outperforms LongT5. Note that {\ours}\textsubscript{base++} have about 290M parameters, while LongT5\textsubscript{large} contains about 770M parameters and LongT5\textsubscript{xl} contains about 3B parameters. From the results, we see that in all the tasks, our base-sized models have on par or better performance compared with LongT5\textsubscript{large}. On the MultiNews dataset, our model even outperforms LongT5\textsubscript{xl}, which is over ten times larger.

\section{Analysis}

\begin{figure*}[t!]
\centering
\begin{minipage}{0.55\textwidth}
    \centering
    \includegraphics[width=0.98\textwidth]{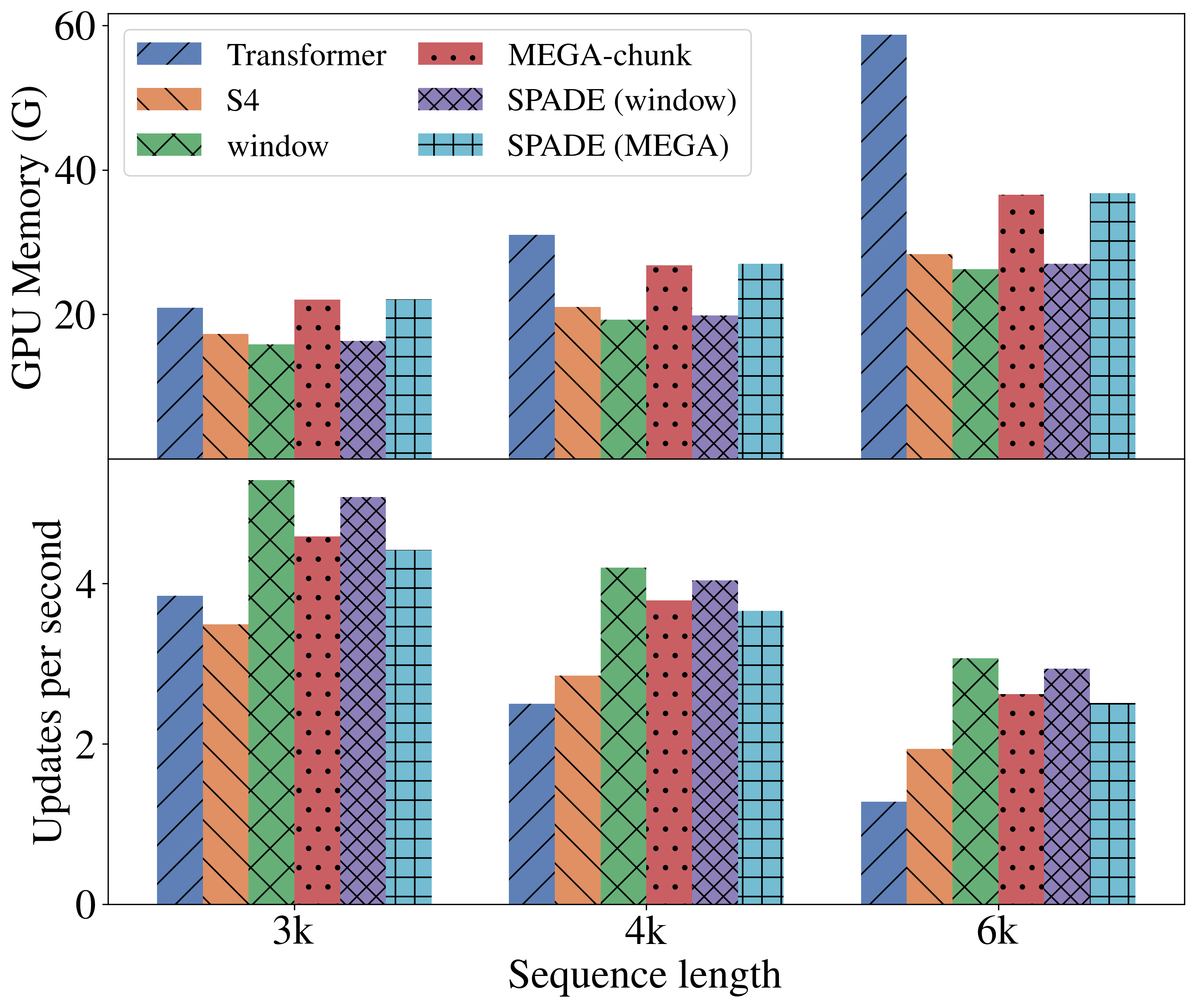}
    \vspace{-0.1in}
    \captionof{figure}{Efficiency comparison on language modeling.}
    \label{fig:efficiency}
\end{minipage} \hfill
\begin{minipage}{0.4\textwidth}
    \centering
    \includegraphics[width=0.9\textwidth]{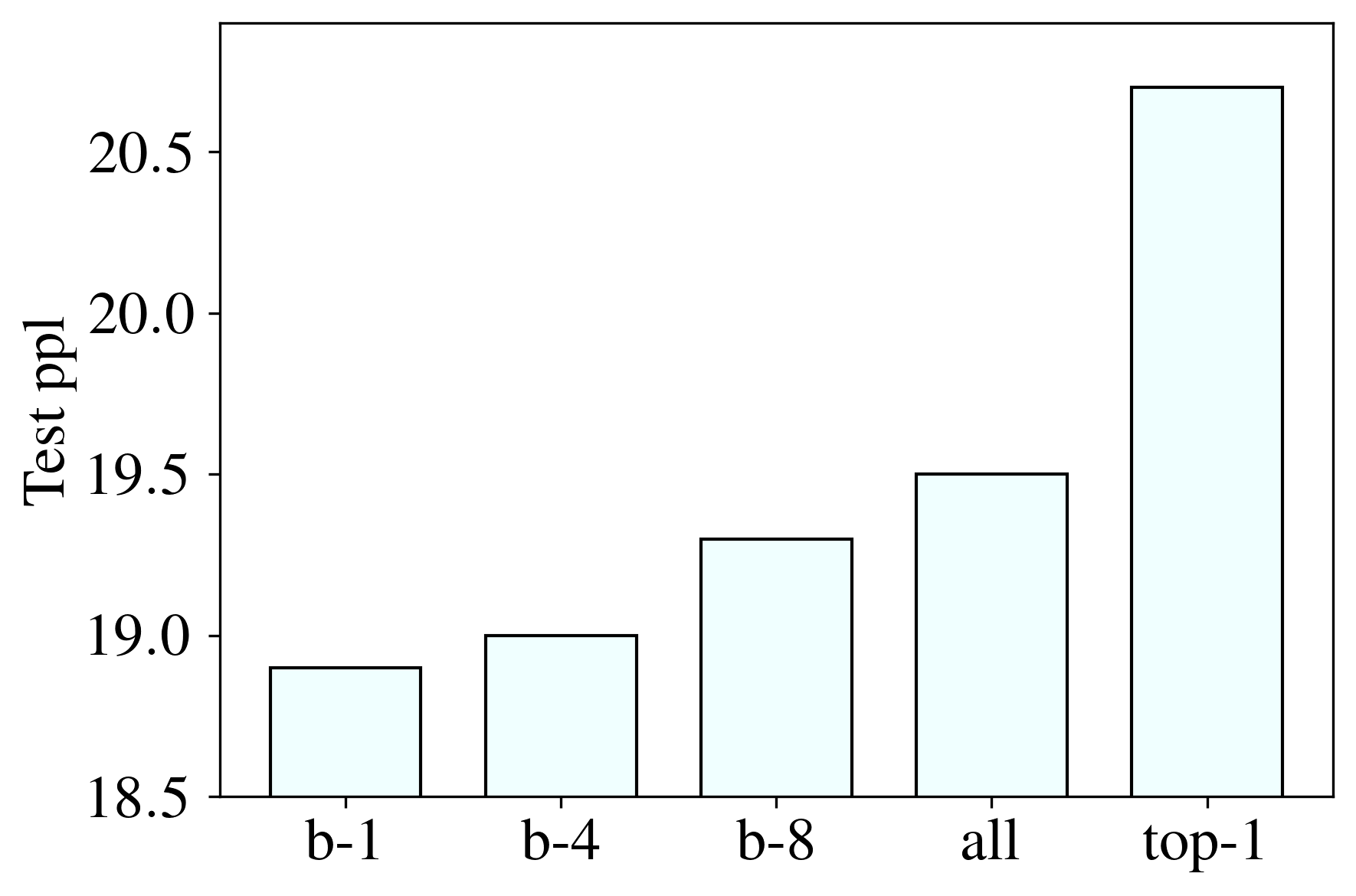}
    \vspace{-0.1in}
    \captionof{figure}{Performance vs. location of SSM. We conduct language modeling experiments with softmax-window attention (window=256). By default, the model has 16 layers, where the bottom layer is a global layer and the rest are local layers. Here, ``\textit{b-k}'' means the bottom-k layers are global layers, ``\textit{all}'' means all layers are global layers, and ``\textit{top-1}'' means the top-1 layer is a global layer.}
    \label{fig:s4-position}
\end{minipage}
\end{figure*}

\newcolumntype{C}{@{\hskip4pt}c@{\hskip4pt}}
\begin{figure*}[t!]
\centering
\begin{minipage}{0.65\textwidth}
    \centering
    \includegraphics[width=0.49\textwidth]{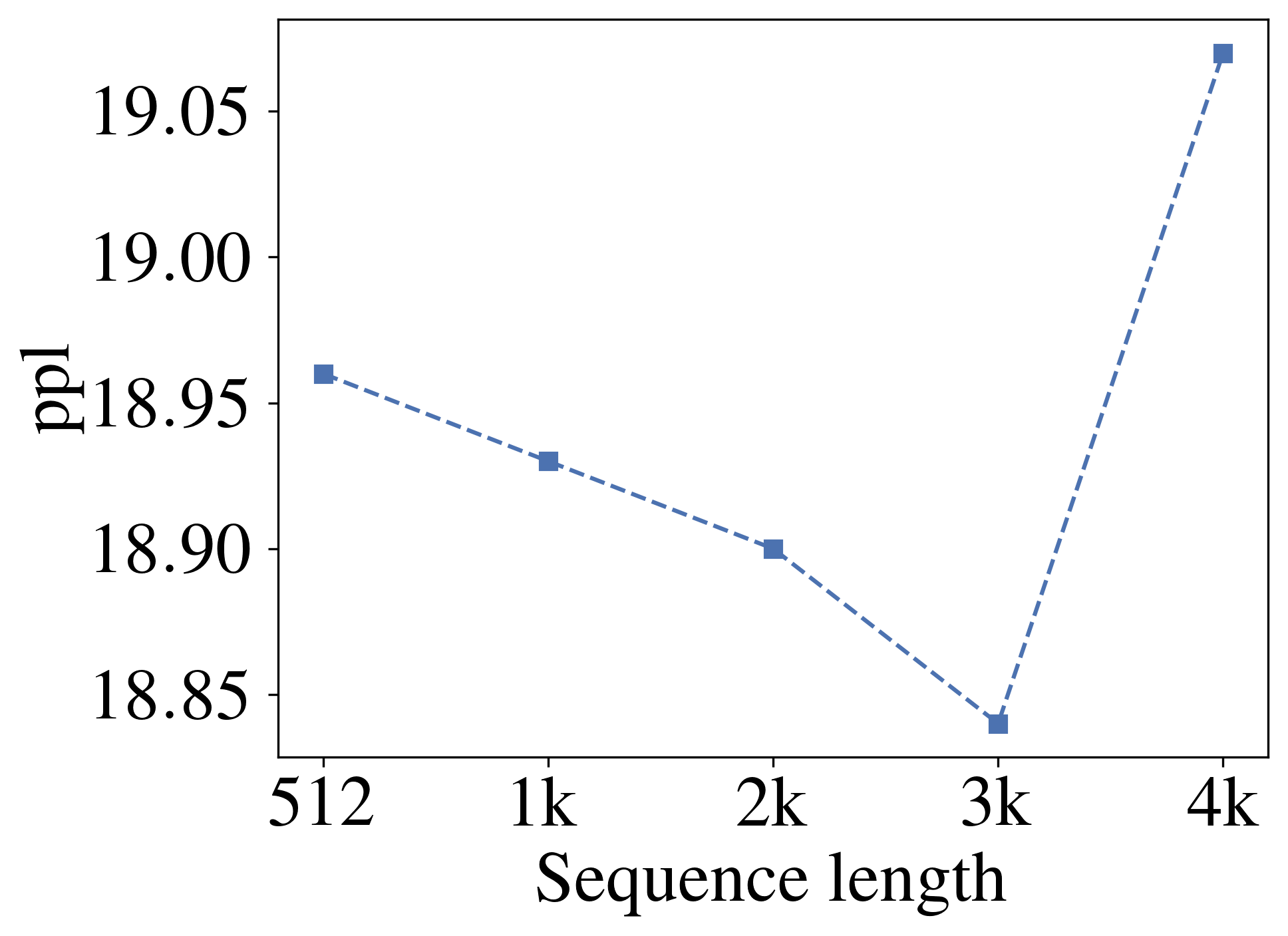}
    \includegraphics[width=0.49\textwidth]{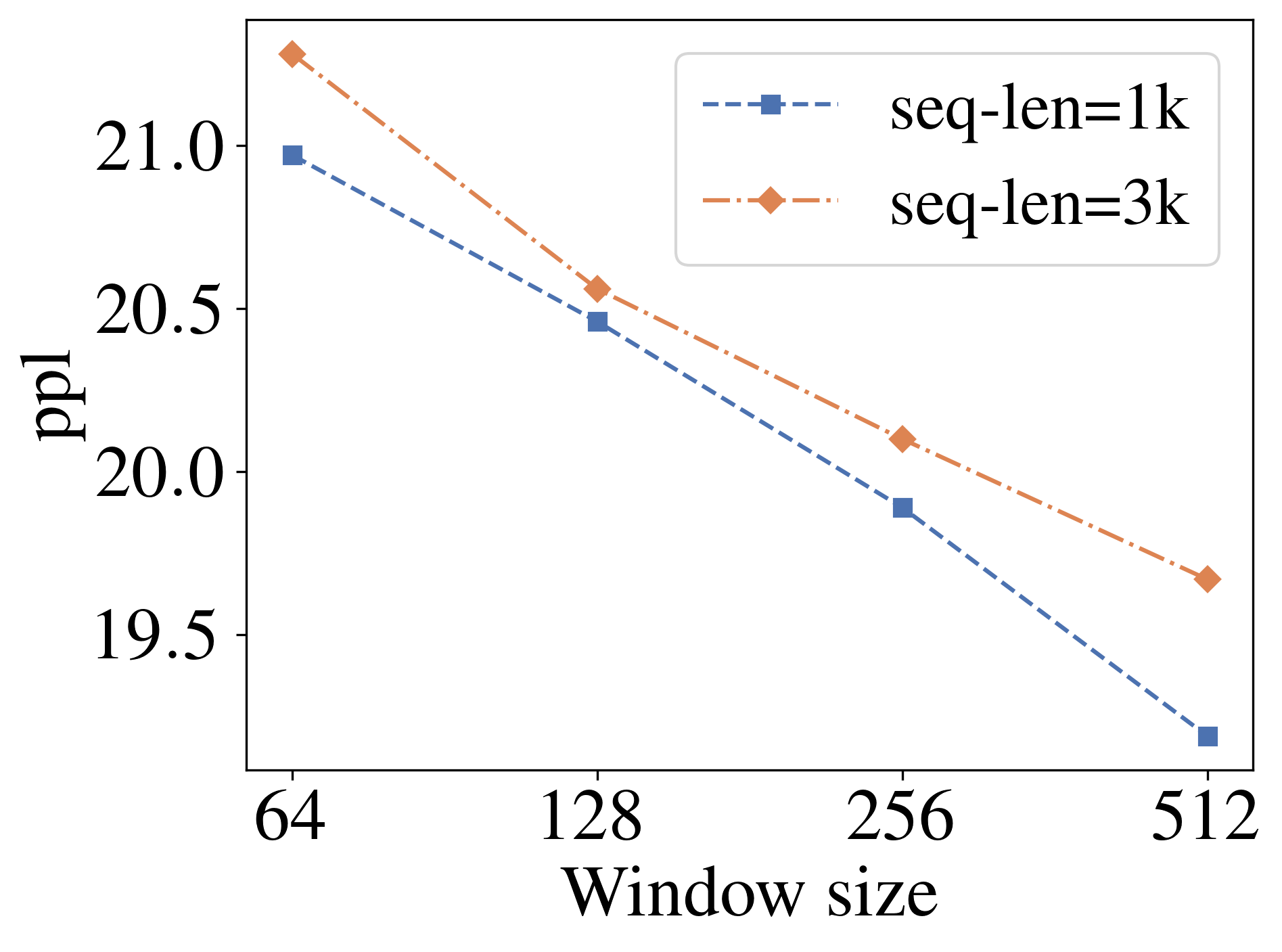}
    \vspace{-0.1in}
    \captionof{figure}{Performance with different configurations. Left: Transformer with full attention using different sequence length; Right: Transformer with window attention using different window size and sequence length.}
    \label{fig:ablation-transformer}
\end{minipage} \hfill
\begin{minipage}{0.32\textwidth}
    \centering \small
    \begin{tabular}{l|CCCC}
        \toprule
        & \multicolumn{4}{c}{\textbf{Sequence length}} \\
        & 2k & 3k & 4k & 6k \\ \midrule
        128 & 19.55 & 19.42 & 19.50 & 19.55 \\
        256 & 18.90 & 18.95 & 18.99 & 19.00 \\
        512 & 18.63 & 18.64 & 18.52 & 18.67 \\
        \bottomrule
    \end{tabular}
    \captionof{table}{Performance of {\ours} with softmax-window using different sequence length and window size. The first column is window size. We conduct language modeling experiments on Wikitext-103.}
    \label{tab:ablation-ours}
\end{minipage}
\end{figure*}

\subsection{Efficiency Comparison}

We compare the efficiency of {\ours} with other models: Transformer with softmax-window attention, Trnasformer with MEGA-chunk attention, and S4. The results are illustrated in Figure~\ref{fig:efficiency}. We see that {\ours} is efficient in terms of both training speed and GPU memory usage. 
For example, when the sequence length is 6k, Transformer uses about 60GB of GPU memory, whereas {\ours} with softmax-window uses only 27GB. Moreover, notice that {\ours} also trains significantly faster than the vanilla Transformer under all settings. Notice that S4 may be less efficient than the vanilla Transformer (e.g., when the sequence length is 3k). This is because in the multi-layer model in \citealt{gu2021efficiently}, each layer contains multiple S4 modules and expensive non-linear components. Therefore, the per-layer computational cost can exceed full attention when the sequence is not extremely long.

We remark that even though we add a S4 module to the bottom layer of {\ours}, such an additional module does not induce much computational overhead. We see that both the training speed and the memory usage of {\ours} with softmax-window is only marginally different from those of window-attention Transformer. We have similar observations for the chunk attention variant.

\subsection{Location and Number of Global Layers}

Recall that in {\ours}, the bottom layer is equipped with a SSM and serves as the global layer, while the rest are local layers (see Figure~\ref{fig:model-arch}). In Figure~\ref{fig:s4-position}, we empirically justify this design choice.

We first investigate the possibility of incorporating more global layers: we set the bottom 1 (the default choice), 4, 8, and 16 (all) layers as global layers. From the results, we see that model performance decreases as we use more global layers. This is because the SSM in the bottom layer captures and filters out global information, such that subsequent SSMs only introduce noise to the intermediate representations.

We also investigate whether the global layer can be the top instead of the bottom layer in {\ours}. From Figure~\ref{fig:s4-position}, we see that model performance drops significantly. This is because as a global information extractor, the global layer encodes positional information, on which the local attention modules rely. Therefore, using the global layer as the top layer is akin to using Transformer models without positional encoding, which will yield unsatisfactory performance.

\subsection{Different Configurations}

We examine how performance changes when we change the sequence length and window size.

From Figure~\ref{fig:ablation-transformer} (left), we see that when we increase the sequence length from 512 to 3k, performance of Transformer with full attention increases. However, when we further increase the sequence length to 4k, model performance drastically drops. This is because in long sequences, the signal-to-noise ratio is low, such that the full attention may easily fit to the noise. From Figure~\ref{fig:ablation-transformer} (right), we see that performance of Transformer with window attention increases when we increase the window size. Moreover, model performance is better with shorter sequences for the same window size. Such findings indicate that performance of window attention depends on the proportion of information within its perception.

From Table~\ref{tab:ablation-ours}, we see that for the same sequence length, performance of {\ours} increases when we increase the window size. Also, we see that performance of {\ours} marginally decreases when we increase the sequence length from 4k to 6k. Recall from Figure~\ref{fig:ablation-transformer} (left) that performance of Transformer with full attention drastically deteriorates when we increase the length from 3k to 4k. Such a result indicates that the proposed model is more suitable to model long sequences.
\section{Related Works}

\subsection{Efficient Transformer Models}

In Eq.~\ref{eq:attention}, we have $\Qb, \Kb, \Vb \in \RR^{L \times d}$, such that computing the attention $\mathrm{Attn}(\Xb)$ introduces $O(L^2)$ time and space costs. Such quadratic costs are prohibitive when the sequence length $L$ is large. There are various attempts to reduce the quadratic time and space complexity of the vanilla attention.

One approach is to employ \textit{sparse attention}. That is, each token only attends to a subset of all the tokens according to pre-defined patterns, e.g., neighboring tokens within a fixed size window. Some examples include Sparse Transformer \citep{child2019generating}, BlockBERT \citep{qiu2019blockwise}, Longformer \citep{beltagy2020longformer}, ETC \citep{ainslie2020etc}, BigBird \citep{zaheer2020big}, HEPOS \citep{huang2021efficient}, and Poolingformer \citep{zhang2021poolingformer}.

Another approach is to use \textit{low-rank projection}. For example, in Linformer \citep{wang2020linformer}, the attention mechanism in Eq.~\ref{eq:attention} becomes 
$\mathrm{Attn}(\Xb) = \mathrm{softmax}(\Qb (\Eb \Kb)^\top / \sqrt{d}) (\Fb \Vb)$. Here, the two additional parameters satisfy $\Eb, \Fb \in \RR^{r \times L}$, where $r$ is the projection rank such that $r \ll L$. Similar methods include Nyströmformer \citep{xiong2021nystromformer}, Synthesizer \citep{tay2021synthesizer}, Transformer-LS \citep{zhu2021long}, and Luna \citep{ma2021luna}.
However, these approaches face difficulty when handling causal tasks, such as auto-regressive language modeling. Specifically, in Eq.~\ref{eq:attention}, we mask out the upper triangular part in the attention score matrix $\Ab \in \RR^{L \times L}$ such that each token can only attend to its previous tokens. However, this is implausible in Linformer since we project the $L\times L$ matrix to a $L \times r$ matrix.

\textit{Kernel-based approaches} can be used to approximate the full attention $\mathrm{Attn}(\Xb)$. In these approaches, the quadratic-time softmax attention is replaced by fast linear-time kernel approximations (e.g., Gaussian and arc-cosine kernel). Some examples include Linear Transformer \citep{katharopoulos2020transformers}, Performer \citep{choromanski2020rethinking}, Random Feature Attention \citep{peng2021random}, and FMMformer \citep{nguyen2021fmmformer}.
Both low-rank projection and kernel-based approaches approximate the full attention, and thus, they often suffer from non-negligible approximation error.

We can also adopt \textit{clustering-based approaches}, where we divide $\Qb$ or $\Kb$ into several clusters, and only perform inter-cluster attention. Such methods include Reformer \citep{kitaev2020reformer}, Clusterformer \citep{wang2020cluster}, Sinkhorn Transformer \citep{tay2020sparse}, Fast Transformer \citep{vyas2020fast}, Routing Transformer \citep{roy2021efficient}, and FLASH \citep{hua2022transformer}.

\subsection{Pre-Trained Language Models}

Pre-trained language models \citep{devlin2018bert, liu2019roberta, raffel2020exploring, brown2020language, he2020deberta} have achieved state-of-the-art performance on various natural language processing tasks. However, most of these models are not suitable for long sequences. For example, BERT \citep{devlin2018bert} uses a fixed-length positional embedding, such that it cannot handle sequences with length more than 512.
In contrast, LongT5 \citep{guo2021longt5} facilitates training on long sequences by leveraging relative positional embedding \citep{shaw2018self} and efficient attention methods. The model targets long sequence modeling tasks such as text summarization.
\section{Conclusion}

In this work, we propose {\ours}, a state space augmented Transformer model that targets long sequence modeling. {\ours} is a multi-layer Transformer model, where the bottom layer is a global layer and the rest are local layers. In the global layer, we use a SSM to augment coarse global information, which are subsequently refined by the following local layers. We instantiate the local layers with off-the-shelf efficient attention methods, such as window attention. The proposed model has linear time and space computationally complexity, facilitating it to handle long sequences.
We conduct extensive experiments on the Long Range Arena (LRA) benchmark and language modeling datasets to demonstrate the effectiveness and efficiency of {\ours}. We also pre-train encoder-decoder models to demonstrate the scalability of {\ours}, and we perform fine-tuning experiments on natural language understanding (GLUE) and natural language generation (summarization) tasks. In all the experiments, {\ours} exhibits superior performance and outperforms the baselines.
\section*{Acknowledgments}
We thank Hao Cheng, Bin Yu, Jianwei Yang, Baolin Peng, Minjia Zhang and Linyuan Gong for valuable discussions and comments, and Microsoft Research Technology Engineering team for setting up GPU machines.


\bibliography{custom}
\bibliographystyle{acl_natbib}

\clearpage
\appendix
\section{Training Details}

\subsection{Language Model Pre-Training and Fine-Tuning}

For language model pre-training and fine-tuning experiments, we use Adam \citep{kingma2014adam} as the optimizer. Hyper-parameters for pre-training are detailed in Table~\ref{tab:hp4pretrain}; and hyper-parameters for fine-tuning are detailed in Table~\ref{tab:hp4ft}.

\begin{table*}[t]
\centering
\begin{tabular}{l*{3}{c}}
\toprule
\textbf{Parameters} & \textbf{Base} & \textbf{Base++}  \\
\midrule
Peak Learning Rate & 4e-4 & 2e-4 \\
Batch Size & 2,048 & 2,048 \\
Warmup Steps & 10,000 & 10,000 \\
Total Steps & 125,000 & 2,000,000 \\
Sequence Length & 1024 & 1024 \\
Relative Position Encoding Buckets & 32 & 32 \\
Relative Position Encoding Max Distance & 128 & 128 \\
Adam $\epsilon$ & 1e-6 & 1e-6 \\
Adam ($\beta_1$, $\beta_2$) & (0.9, 0.98) & (0.9, 0.98)\\
Clip Norm & -- & 1.0 \\
Dropout & 0.1 & 0.1  \\
Weight Decay & 0.01 & 0.01 \\
\bottomrule
\end{tabular}
\vskip -0.05in
\caption{Hyper-parameters for pre-training.}
\label{tab:hp4pretrain}
\end{table*}

\begin{table*}[t]
\centering
\begin{tabular}{lcc}
\toprule
\textbf{Parameters} & \textbf{Range} \\ \midrule
Learning Rate & \{2e-5, 4e-5, 5e-5, 1e-4\} \\
Batch Size & \{16, 32\}  \\
Maximum Training Epochs & \{3, 5, 10\}  \\
Dropout & 0.1   \\
Warmup Step Rate & 0.1 \\
Weight Decay & 0.1  \\
\bottomrule
\end{tabular}
\vskip -0.05in
\caption{Hyper-parameters for fine-tuning.}
\label{tab:hp4ft}
\end{table*}

\begin{table*}[t]
\centering
\begin{tabular}{lcc}
\toprule
\textbf{Parameters} & \textbf{Range} \\ \midrule
Learning Rate & \{1e-3, 3e-3, 5e-3, 1e-2\} \\
Dropout & \{0.0, 0.1, 0.2\}   \\
Weight Decay & \{0.0, 0.01, 0.03, 0.05\}  \\
\bottomrule
\end{tabular}
\vskip -0.05in
\caption{Hyper-parameters for training on LRA.}
\label{tab:hp4lra}
\end{table*}

\subsection{Long Range Arena}
We follow the model architecture settings in \citealt{ma2022mega}. The rest of the hyper-parameters are detailed in Table~\ref{tab:hp4lra}.

\subsection{Language Modeling}
We follow the settings in \citealt{baevski2018adaptive}, including model architecture and hyper-parameters.

\section{Dataset Details}
\label{app:dataset}

Statistics of the GLUE benchmark is summarized in Table~\ref{tab:glue}. 

\begin{table*}[t!]
\centering
\begin{tabular}{l|l|c|c|c|c|c}
\toprule 
\bf Corpus & Task & \# Train & \# Dev & \# Test  & \# Labels & Metrics\\ \midrule
\multicolumn{6}{@{\hskip1pt}r@{\hskip1pt}}{Single-Sentence Classification} \\ \hline
CoLA & Acceptability&8.5k & 1k & 1k & 2 & Matthews corr\\ \hline
SST & Sentiment&67k & 872 & 1.8k & 2 & Accuracy\\ \midrule
\multicolumn{6}{@{\hskip1pt}r@{\hskip1pt}}{Pairwise Text Classification} \\ \hline
MNLI & NLI& 393k& 20k & 20k& 3 & Accuracy\\ \hline
RTE & NLI &2.5k & 276 & 3k & 2 & Accuracy \\ \hline
QQP & Paraphrase&364k & 40k & 391k& 2 & Accuracy/F1\\ \hline
MRPC & Paraphrase &3.7k & 408 & 1.7k& 2&Accuracy/F1\\ \hline
QNLI & QA/NLI& 108k &5.7k&5.7k&2& Accuracy\\ \midrule
\multicolumn{5}{@{\hskip1pt}r@{\hskip1pt}}{Text Similarity} \\ \hline
STS-B & Similarity &7k &1.5k& 1.4k &1 & Pearson/Spearman corr\\ \bottomrule
\end{tabular}
\vskip -0.05in
\caption{Statistics of the GLUE benchmark.}
\label{tab:glue}
\end{table*}

\end{document}